\documentclass[lettersize,journal]{IEEEtran}

\usepackage{amsmath,amssymb,amsfonts}
\usepackage{textcomp}

\usepackage{graphicx}
\usepackage{tikz}
\usepackage{xcolor}

\usepackage{subcaption} 

\usepackage{booktabs}
\usepackage{multirow}
\usepackage{array}
\usepackage{tabularx}

\usepackage{algorithmic}
\usepackage{stfloats}
\usepackage{url}
\usepackage{verbatim}
\usepackage{comment}
\usepackage{afterpage}
\usepackage{etoolbox}
\usepackage{cite}

\def\BibTeX{{\rm B\kern-.05em{\sc i\kern-.025em b}\kern-.08em
    T\kern-.1667em\lower.7ex\hbox{E}\kern-.125emX}}

\begin{document}
\title{ReaLiTy and LADS: A Unified Framework and Dataset Suite for LiDAR Adaptation Across Sensors and Adverse Weather Conditions}
\author{Vivek Anand, Bharat Lohani, Rakesh Mishra, and Gaurav Pandey 
\thanks{Vivek Anand and Bharat Lohani are with the Geoinformatics Group, Department of Civil Engineering, Indian Institute of Technology Kanpur, India (email: viveka21@iitk.ac.in, blohani@iitk.ac.in)}%
\thanks{Rakesh Mishra is with the Department of Geodesy and Geomatics Engineering, University of New Brunswick, NB, Canada (email: rakesh.mishra@unb.ca)}%
\thanks{Gaurav Pandey is with Department of Engineering Technology \& Industrial Distribution (ETID) of Texas A\&M University (email: gpandey@tamu.edu)}%
}
\maketitle

\begin{abstract}
Reliable LiDAR perception requires robustness across sensors, environments, and adverse weather. However, existing datasets rarely provide physically consistent observations of the same scene under varying sensor configurations and weather conditions, limiting systematic analysis of domain shifts. This work presents \textit{ReaLiTy}, a unified physics-informed framework that transforms LiDAR data to match target sensor specifications and weather conditions. The framework integrates physically grounded cues with a learning-based module to generate realistic intensity patterns, while a physics-based weather model introduces consistent geometric and radiometric degradations. Building on this framework, we introduce the \textit{LiDAR Adaptation Dataset Suite (LADS)}, a collection of physically consistent, transformation-ready point clouds with one-to-one correspondence to original datasets.  Experiments demonstrate improved cross-domain consistency and realistic weather effects. ReaLiTy and LADS provide a reproducible foundation for studying LiDAR adaptation and simulation-driven perception in intelligent transportation systems.
\end{abstract}


\section{Introduction}

Autonomous vehicles (AVs) rely on accurate perception to operate safely in complex environments. Among sensing modalities, LiDAR plays a central role by providing high-resolution 3D geometry independent of lighting conditions, supporting tasks such as object detection, mapping, and localization \cite{wang2018pointseg}.

Collecting large-scale real-world data across diverse sensors and weather conditions is expensive and time-consuming. Simulation has therefore become essential for scalable data generation and safe testing \cite{anand2024toward}. However, its effectiveness depends on how well synthetic data reflects real sensor behavior and environmental effects \cite{anand2026siat}.

Realistic LiDAR simulation remains challenging due to the combined influence of sensor characteristics, scene geometry, material properties, and atmospheric conditions \cite{hahner2022lidar}. Physics-based methods offer interpretability but struggle with complex sensor responses and weather interactions, while learning-based approaches capture nonlinearities but often lack physical consistency and generalization across domains \cite{vacek2021learning, Vivek_Advancing}. As a result, existing solutions remain fragmented and require significant adaptation for new sensors and conditions.

LiDAR intensity, which reflects the strength of returned signals, is critical for perception tasks but varies significantly across sensors and environments. It depends on physical factors such as range and reflectance, as well as weather-induced effects including attenuation, scattering, and noise. Sensor-specific characteristics further introduce variability, causing the same scene to produce different observations under different configurations and conditions.

A key challenge is the joint modeling of cross-sensor variability and weather-induced degradation. Existing approaches typically address these factors independently and lack a unified, physically grounded framework. Progress is further limited by the absence of paired datasets capturing the same scene under varying sensors and weather, restricting systematic evaluation and reproducibility.

To address these challenges, we introduce a unified physics-informed learning framework that transforms LiDAR data to target sensor characteristics and adverse-weather conditions. We further release a derived dataset suite that provides physically consistent, transformation-ready point clouds with one-to-one correspondence to original datasets. Together, these contributions enable scalable and reproducible research on LiDAR adaptation and simulation-driven perception.

\textbf{Contributions:}
\begin{itemize}
    \item \textbf{ReaLiTy: Realistic LiDAR Transformation Framework} for sensor adaptation and weather-aware LiDAR transformation.
    \item \textbf{LADS: LiDAR Adaptation Dataset Suite} with physically consistent, transformation-ready point clouds across sensors and weather conditions.
    \item \textbf{Benchmarks and Pretrained Models} enabling standardized and reproducible evaluation.
\end{itemize}

All framework details and the full dataset suite are available on the \textit{Project page:} \url{https://voodooed.github.io/ReaLiTy/} and also integrated with the SimDaaS simulator \cite{simdaas}.

\section{Related Work}

\subsection{LiDAR Simulation and Intensity Modeling}

Physics-based simulators model LiDAR returns using geometric optics and material interactions. Prior works combine ray casting with scene priors to generate realistic geometry and intensity, sometimes augmented with data-driven corrections to better match real sensors \cite{manivasagam2020lidarsim, guillard2022learning}. While these approaches offer physical interpretability, they often struggle to capture complex sensor-specific responses and environmental distortions.

Learning-based methods instead use neural networks to map simulated data to real sensor behavior, achieving strong perceptual realism \cite{learning-based-review}. However, these approaches typically lack explicit physical constraints, limiting generalization across sensors and unseen conditions.

\subsection{Weather Effects and Domain Adaptation}

Adverse weather introduces attenuation, scattering, and noise that significantly affect LiDAR measurements. Existing works model such effects using physical formulations or empirical datasets \cite{bijelic2020weather, zhang2023fogsimulation}. Domain adaptation methods further address cross-sensor and environmental shifts using learning-based techniques \cite{fang2023lidardiffusion}. However, most approaches treat sensor variation and weather effects independently and focus on downstream perception tasks rather than raw LiDAR intensity modeling.

\subsection{Datasets and Limitations}

Large-scale datasets such as SemanticKITTI, nuScenes, and KITTI have advanced LiDAR perception research \cite{behley2019iccv, caesar2020nuscenes, geiger2012we}, while others capture adverse weather scenarios \cite{mclean2020cadc, bijelic2020weather}. However, these datasets are not paired across sensors or weather conditions, limiting systematic evaluation of domain shifts.

Simulation-based datasets provide controlled variations but are typically limited to RGB data or lack realistic LiDAR intensity modeling. Moreover, existing resources do not offer paired LiDAR observations reflecting both sensor-specific characteristics and weather-induced degradation.

\subsection{Summary}

Existing methods remain fragmented across physics-based modeling, learning-based simulation, weather modeling, and domain adaptation. Furthermore, the absence of paired datasets across sensors and weather conditions limits reproducibility and benchmarking. These gaps motivate the development of a unified framework and dataset suite for LiDAR sensor and weather adaptation.

\section{Realistic LiDAR Transformation (ReaLiTy) Framework}

\subsection{Problem Formulation}

LiDAR intensity varies across sensors and environmental conditions due to differences in hardware, material interactions, and atmospheric effects. Let $\mathcal{D}_s$ denote a source domain (sensor or weather condition) and $\mathcal{D}_t$ a target domain. The goal is to learn a transformation
\[
\mathcal{F}: \mathcal{D}_s \rightarrow \mathcal{D}_t
\]
that maps source-domain LiDAR data to match the physical and statistical characteristics of the target domain.

\subsection{Modular Architecture Overview}

\begin{figure*}[t]
\centering
\includegraphics[width=0.8\linewidth]{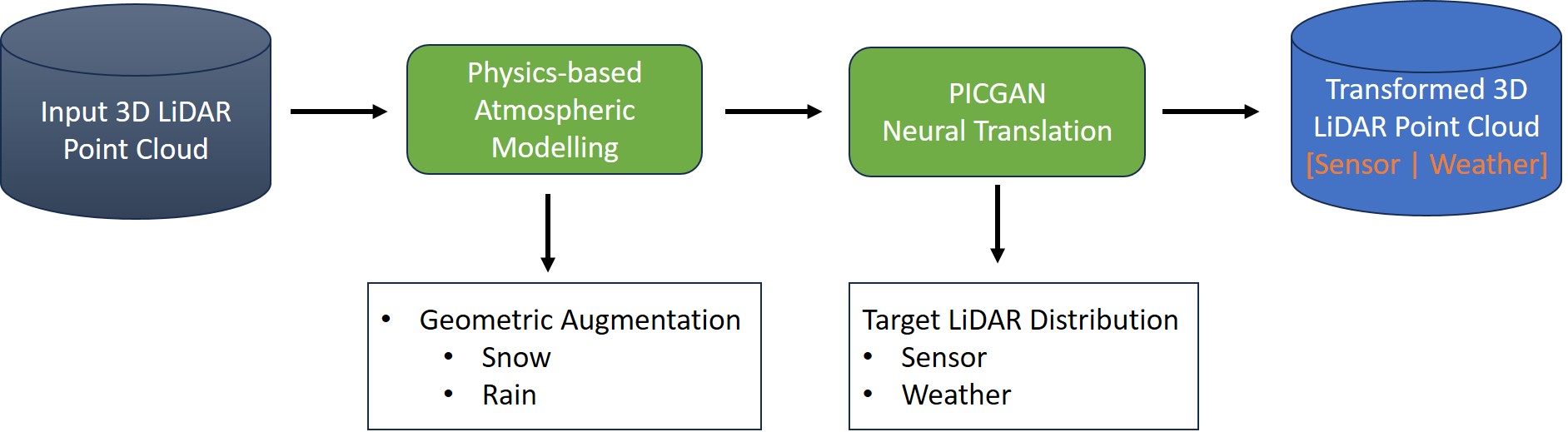}
\caption{Overview of the proposed ReaLiTy framework. A modular design with physics-guided sensor and weather adaptation branches driven by shared physical cues such as range, incidence angle, and material reflectance.}
\label{fig:ReaLiTy_framework}
\end{figure*}

The proposed framework adopts a modular architecture that unifies sensor and weather adaptation under a shared physics-guided learning paradigm (Fig.~\ref{fig:ReaLiTy_framework}). Both branches rely on common physical cues—range, incidence angle, and material reflectance—to ensure physically consistent intensity transformations.

\subsubsection{Sensor Adaptation Branch}

In sensor adaptation, variation primarily appears in intensity rather than geometry. This stage focuses on transforming intensity distributions to match target sensor characteristics while preserving point geometry. The mapping accounts for differences in channel layout, beam divergence, emitted energy, reflectance calibration, and internal signal processing.

\subsubsection{Weather Adaptation Branch}

Weather affects both intensity and geometry. Rain and snow introduce attenuation, scattering, backscatter, and noise, leading to structural distortions such as point dropouts and spurious returns. To model these effects, we incorporate a physics-based weather module \cite{lisa_kilic2021lidar} that converts clear-weather sweeps into adverse-weather counterparts via particle–beam interactions and noise processes.

The resulting point clouds are then processed by the learning framework for intensity prediction and refinement using physically meaningful inputs such as range, incidence angle, and reflectance \cite{anand_snow_iv, Vivek_Advancing}. Geometry changes are handled by the physics module, while intensity is estimated by the learning module. In contrast, sensor adaptation typically preserves geometry and modifies only intensity.

\subsubsection{Unified Learning Principle}

Both branches share a unified physics-guided learning strategy combining adversarial, consistency, and physics-based losses. Physical cues anchor the learning process, while adversarial objectives align intensity distributions with the target domain and consistency constraints preserve structural content. A physics-based loss further enforces plausible signal behavior, and data-driven components capture complex sensor and environmental effects.

The framework supports sensor-only, weather-only, or joint adaptation within a single architecture. A shared preprocessing stage converts point clouds into range, incidence-angle, and reflectance representations. The system is config-driven and supports pretraining, fine-tuning, and extension to new datasets. The released implementation includes pretrained models, dataset configurations, and reproducibility resources, and serves as the basis for generating the LADS dataset.


\section{LiDAR Adaptation Dataset Suite (LADS)}

A major limitation in LiDAR adaptation research is the absence of a unified dataset providing physically consistent reference outputs for evaluating sensor and weather transformations. Existing datasets contain high-quality LiDAR scans across sensors and environments, but lack systematically transformed counterparts generated under a consistent framework, limiting controlled and reproducible benchmarking.

To address this, we introduce the \textit{LiDAR Adaptation Dataset Suite (LADS)}, a derived dataset generated using the ReaLiTy framework \cite{anand_snow_iv, Vivek_Advancing}. LADS contains transformed point clouds rather than raw data. Each sample corresponds one-to-one with a frame from its source dataset and preserves original file naming, indexing, and directory structure, enabling users to directly construct paired datasets for evaluation.

LADS provides physically consistent targets under two settings: sensor adaptation and weather adaptation. It is built on widely used benchmarks, including SemanticKITTI, nuScenes, and Voxelscape (Table \ref{tab:data_release_summary}). For SemanticKITTI and Voxelscape, sequences 00--10 are included, while for nuScenes, all LiDAR keyframes with lidarseg annotations are provided. For each frame, adverse-weather variants (snow and rain) are released. Additionally, Voxelscape includes intensity-adapted versions aligned with SemanticKITTI and nuScenes, enabling cross-dataset studies.

All transformed point clouds retain the exact format, field ordering, frame indices, and directory hierarchy of their source datasets, allowing direct reuse of calibration, poses, and labels. This ensures seamless integration into existing pipelines and simplifies reproducibility.

LADS currently includes the datasets described above and will be expanded to cover additional sensors, environments, and weather conditions.

\setlength{\tabcolsep}{3pt}
\begin{table}[ht]
\centering
\caption{Released LADS adverse-weather LiDAR data and corresponding source frames. The original file and folder structure is preserved for direct reuse of calibration and label files.}
\renewcommand{\arraystretch}{1.15}
\small
\begin{tabular}{p{2.6cm} p{3.1cm} p{2.5cm}}
\toprule
Source & Frames Used & Variants Provided \\
\midrule
SemanticKITTI 
& Sequences 00--10 
& Snow, Rain  \\

nuScenes lidarseg 
& All labeled keyframes 
& Snow, Rain  \\

Voxelscape 
& Sequences 00--10 
& Snow, Rain, 

Intensity-adapted \\
\bottomrule
\end{tabular}
\label{tab:data_release_summary}
\end{table}

LADS serves as a standardized benchmark for LiDAR adaptation, enabling reproducible evaluation of sensor-transfer and weather-transfer methods, including intensity accuracy, weather realism, geometric fidelity, and downstream perception impact. While not a replacement for original datasets, it provides a unified, physically grounded reference for controlled benchmarking.


\subsection{Licensing and Access}

LADS is released under a permissive academic research license and includes only transformed point clouds. Since raw data belongs to existing datasets, users are directed to official repositories for original scans. All processing scripts and data generation utilities will be publicly released to ensure transparency and reproducibility.

\section{Methodology}

The proposed ReaLiTy framework extends our earlier physics-informed learning architecture \cite{anand2026sim2real} \cite{anand_snow_iv} \cite{anand2026sim2real_aw} \cite{Vivek_Advancing}  by introducing a unified formulation that jointly supports sensor adaptation and weather adaptation within a single, coherent learning pipeline. While it builds on the same core philosophy of physics-informed generative modelling, this work introduces a modular architecture, physics-based weather augmentation, unified multi-domain losses, and a scalable training strategy that handles diverse sensors and environmental conditions. 
For completeness and reproducibility, we summarize the key components below while referring to our earlier work for foundational details where appropriate.

\subsection{Sensor Adaptation Model}

For sensor-to-sensor intensity translation, we employ our physics-informed cycle-consistent GAN (PICGAN) architecture \cite{anand_snow_iv}, redesigned to operate as a plug-in module within a unified framework. The architecture receives the physics-modality stack consisting of range, incidence angle, material reflectance, and the target-domain intensity distribution. It predicts a target-sensor–consistent intensity distribution.

Two generators ($G_{S2T}$ and $G_{T2S}$) and two discriminators ($D_T$ and $D_S$) enable unpaired translation. The forward generator learns the sensor adaptation mapping, while the backward generator enforces invertibility through cycle consistency. 

\subsection{Weather Adaptation Model}

Weather adaptation introduces additional complexity because adverse conditions affect both geometry and intensity. Before learning the intensity mapping, we apply a physics-based geometric augmentation module to convert clear-weather point clouds into noisy, weather-specific observations.

\subsubsection*{Physics-Based Geometric Distortion}
We employ a state-of-the-art physics-based atmospheric scattering model~\cite{lisa_kilic2021lidar} to simulate scatter-induced noise points, range-dependent attenuation, and weather-related point dropouts in rain and snow. The model adopts a hybrid Monte Carlo formulation incorporating Mie scattering theory to estimate extinction coefficients, particle backscatter, and detection probabilities. This module produces weather-realistic geometries that serve as input to the physics-informed learning model. We refer readers to~\cite{lisa_kilic2021lidar} for physics details and provide a concise summary here to maintain reproducibility.

\subsubsection*{Physics-Informed Learning-Based Weather Intensity Transformation}
Given the distorted geometry, the weather generator $G_{C2W}$ transforms clear-weather intensities into weather-specific intensities. A conditioning signal $w \in \{\text{rain}, \text{snow}\}$ is optionally appended to the input channels for multi-style adaptation. The same physics-modality stack (range, incidence, reflectance) is used to guide the mapping.

To support multiple weather styles, we include a consistency constraint ensuring that each weather type preserves its distinct attenuation–backscatter structure while sharing common physical behavior. This multi-style formulation forms a key contribution over earlier work.

\subsection{Loss Functions}

Both adaptation branches are trained under a unified objective composed of adversarial, reconstruction, and physics-guided losses.

\paragraph*{Adversarial Loss}
For a generator $G$ and discriminator $D$, we use the standard GAN loss
\begin{equation}
    \mathcal{L}_{adv}(G, D) 
    = \mathbb{E}_{x \sim p_T}[\log D(x)] +
      \mathbb{E}_{y \sim p_S}[\log(1 - D(G(y)))]
\end{equation}

\paragraph*{Cycle-Consistency Loss}
Cycle consistency encourages invertibility:
\begin{equation}
\begin{aligned}
\mathcal{L}_{cycle} &= 
\mathbb{E}_{x \sim p_T}\left\| x - G_{S2T}(G_{T2S}(x)) \right\|_1 \\
&\quad + 
\mathbb{E}_{y \sim p_S}\left\| y - G_{T2S}(G_{S2T}(y)) \right\|_1
\end{aligned}
\end{equation}

\paragraph*{Physics Consistency Loss}
Following our physics-informed formulation, we anchor the predicted intensity to a physically grounded reference model. 
For clear-weather (sensor adaptation) settings, the physics-based intensity is computed as
\[
I_{\text{phy}} = \frac{MR \cdot \cos(\theta)}{R},
\]
where $MR$ denotes material reflectance, $\theta$ is the incidence angle, and $R$ is the range.

For adverse-weather conditions, atmospheric attenuation is incorporated using the Beer–Lambert law, yielding
\[
I_{\text{AW}} = I_{\text{phy}} \cdot e^{-2\alpha R},
\]
where $\alpha$ represents the weather-dependent attenuation coefficient.

The physics consistency loss is then defined as
\begin{equation}
    \mathcal{L}_{phy} = 
    \mathbb{E}\left[\,| I_{\text{pred}} - I_{\text{P}} |\,\right],
\end{equation}
where $I_{\text{P}}$ corresponds to $I_{\text{phy}}$ for sensor adaptation and $I_{\text{AW}}$ for weather adaptation. 
This constraint stabilizes training and ensures that predicted intensities remain physically plausible across both sensor and weather transformations.

\paragraph*{Final Objective}
\begin{equation}
\mathcal{L}_{total} =
\mathcal{L}_{adv} 
+ \lambda_{cycle}\mathcal{L}_{cycle} 
+ \lambda_{phy}\mathcal{L}_{phy}
\end{equation}

\subsection{Training Setup}

We train all models using Adam with learning rate $1\times10^{-5}$ and momentum parameters $(\beta_1, \beta_2) = (0.5, 0.999)$. All geometries are projected to a $64 \times 1048$ spherical representation for 64-channel data and resized appropriately for 32-channel data. Range, incidence, reflectance, and intensity values are normalized to $[0,1]$. Weather labels, when used, are encoded as one-hot maps broadcast to the spatial resolution. Training and evaluation are conducted on NVIDIA A100 GPUs using PyTorch via  Digital Research Alliance of Canada HPC clusters.

\section{Results and Analysis}

\begin{figure*}[!t]
    \centering
    
    \includegraphics[width=0.9\linewidth, height=5cm]{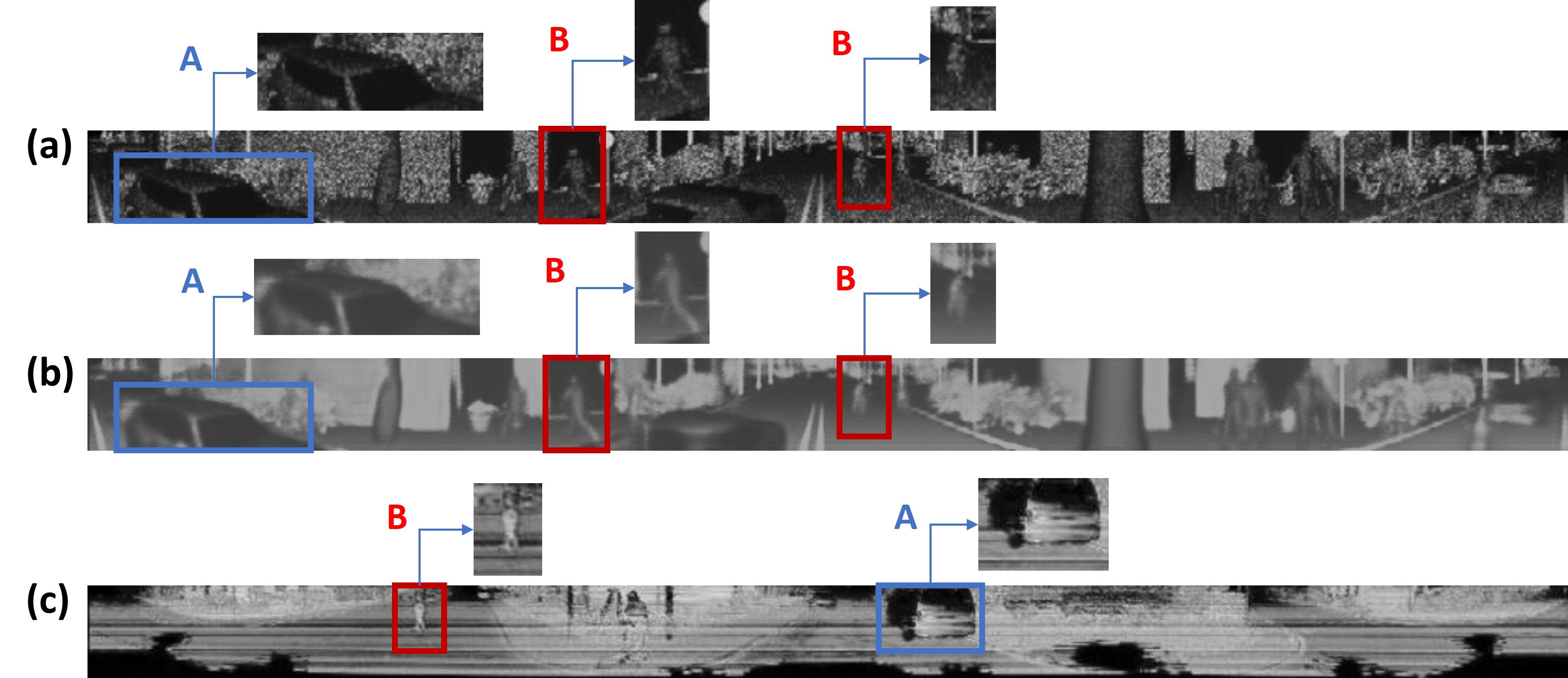}
    
    \caption{\textbf{Qualitative Analysis:} Visual assessment of results from our ReaLiTy framework trained on the SemanticKITTI dataset. a) Physics-modelled simulated LiDAR intensity; b) LiDAR intensity produced by our PICGAN architecture with domain adaptation from the SemanticKITTI dataset; c) Sample LiDAR intensity from the SemanticKITTI dataset. Specific objects are annotated to aid visual comparison: A depicts cars, B depicts pedestrians, C depicts lane markings, and D depicts poles.}
    \label{fig:gi_kitti}
\end{figure*}

\begin{figure*}[!t]
    \centering
    
    \includegraphics[width=0.9\linewidth, height=5cm]{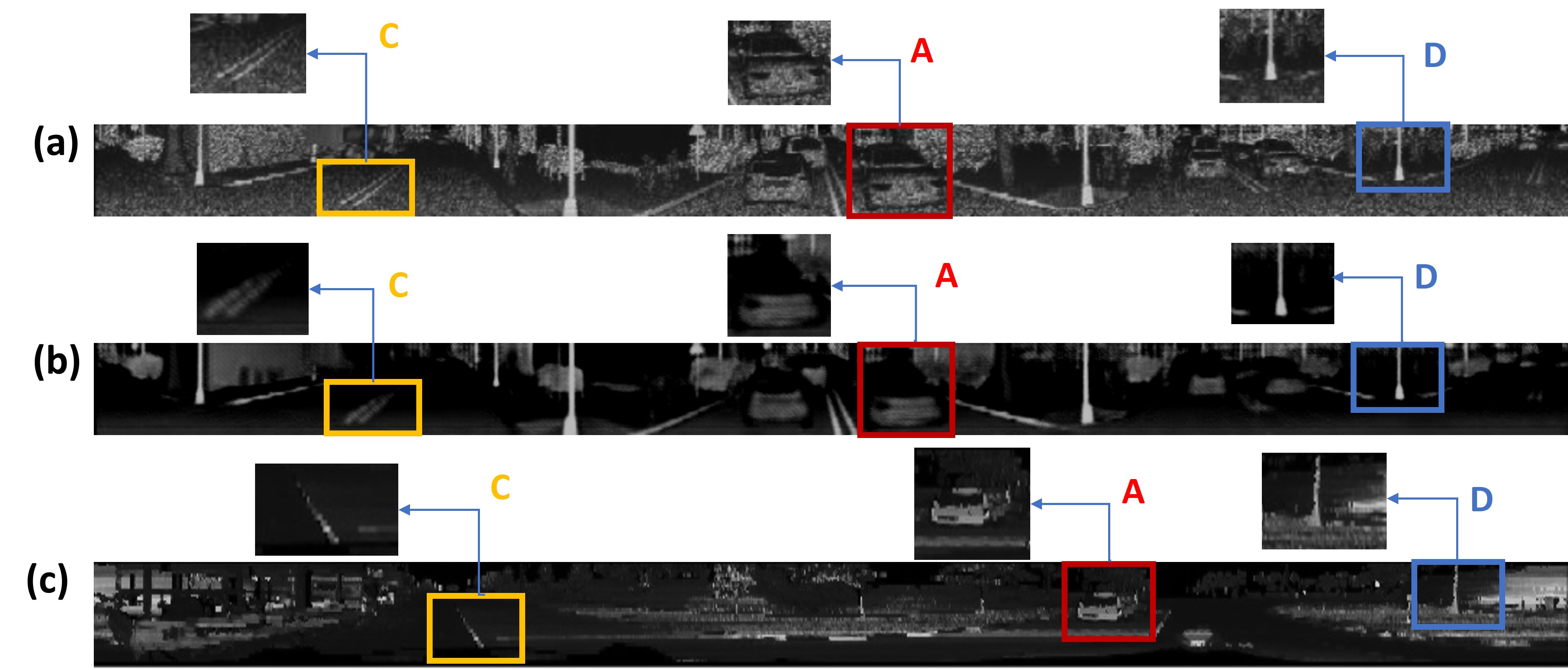}
    
    \caption{\textbf{Qualitative Analysis:} Visual assessment of results from our ReaLiTy framework trained on the nuScenes dataset. a) Physics-modelled LiDAR intensity image; b) LiDAR intensity image produced by our PICGAN architecture with domain adaptation from nuScenes dataset; c) Sample LiDAR intensity image from nuScenes dataset. Specific objects are annotated to aid visual comparison: A depicts cars, B depicts pedestrians, C depicts lane markings, and D depicts poles.}
    \label{fig:gi_nuScenes}
\end{figure*}

\begin{figure*}[!t]
    \centering
    
    \includegraphics[width=0.9\linewidth, height=5cm]{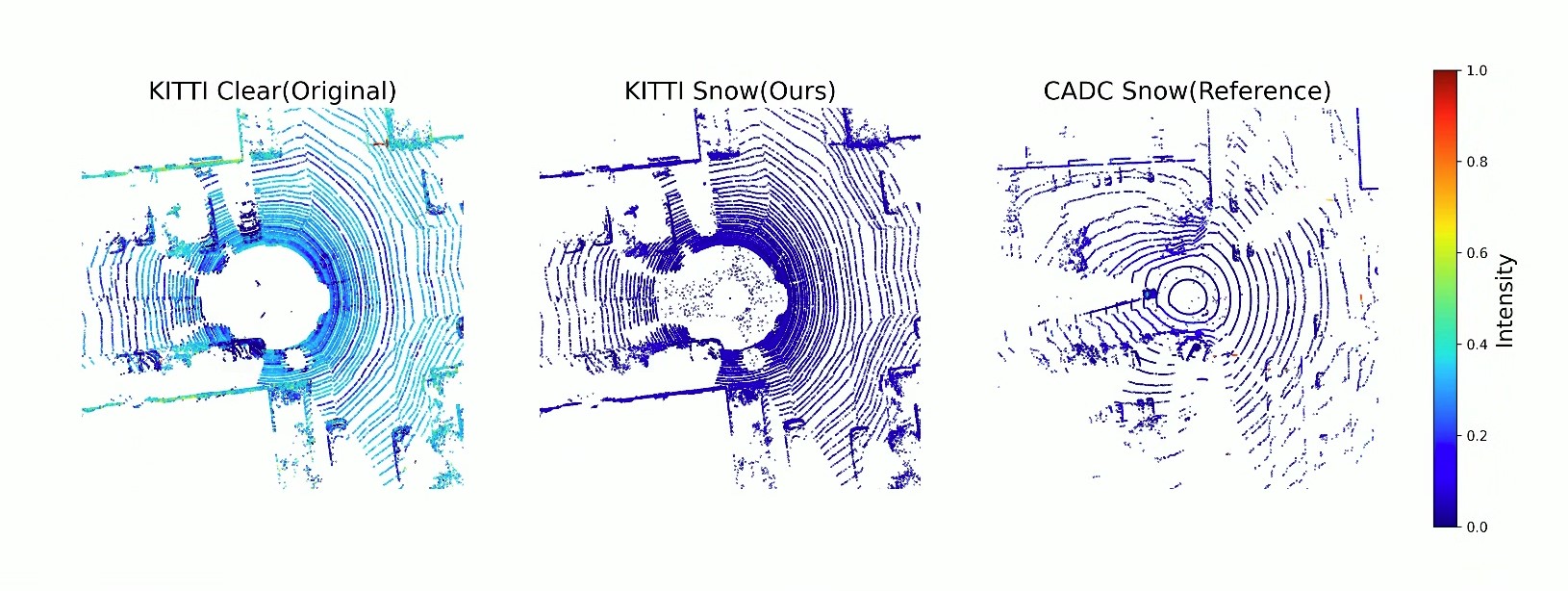}
    
    \caption{\textbf{Qualitative Analysis:} Visual assessment of results from our ReaLiTy framework for snow weather. a) Original KITTI Clear weather data; b) KITTI-Snow weather data produced by our ReaLiTy Framework; c) Reference snow weather data from the CADC dataset.}
    \label{fig:kitti_snow}
\end{figure*}

\begin{figure*}[!t]
    \centering
    
    \includegraphics[width=0.9\linewidth, height=5cm]{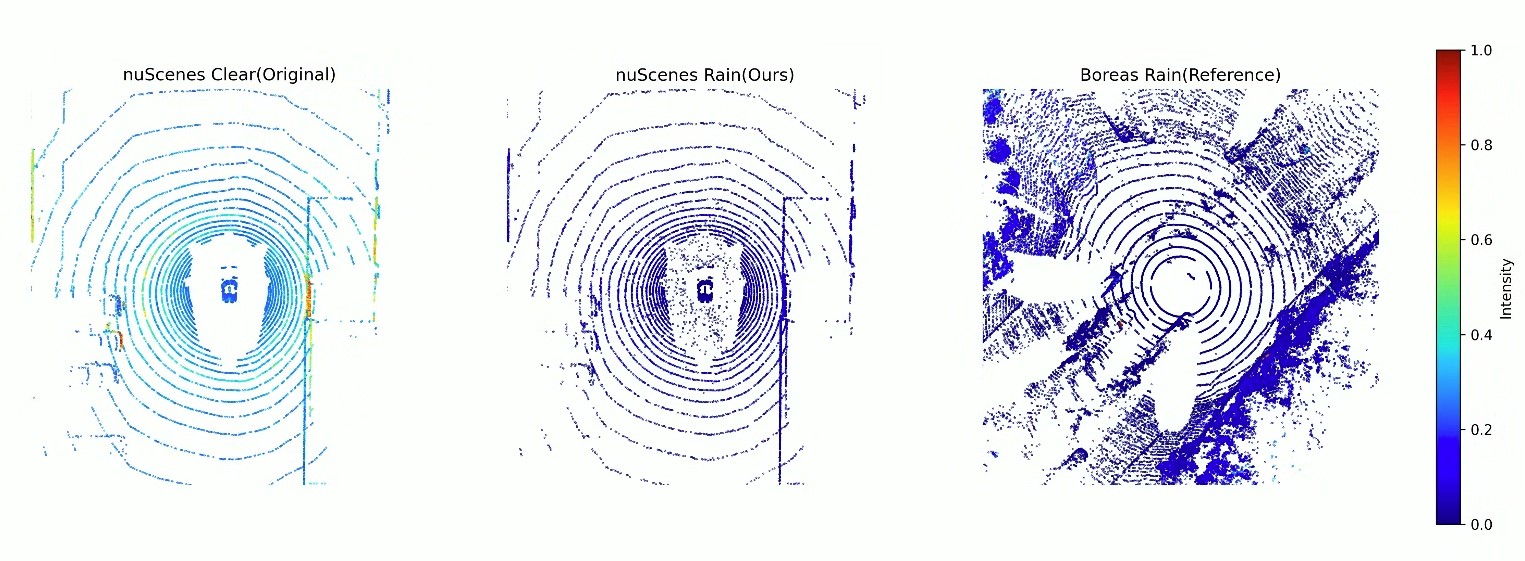}
    
    \caption{\textbf{Qualitative Analysis:} Visual assessment of results from our ReaLiTy framework for rain weather. a) Original nuScenes Clear weather data; b) nuScenes-Rain weather data produced by our ReaLiTy framework; c) Reference rain weather data from the Boreas dataset.}
    \label{fig:nuscenes_rain}
\end{figure*}

This section presents a detailed evaluation of the proposed unified ReaLiTy framework and the LADS weather–adaptation dataset. We report qualitative results across sensor domains and weather domains (snow and rain), followed by quantitative performance measured through a downstream 3D object detection task. All results are reported for both SemantiCKITTI and nuScenes, with real-world CADC (snow) and Boreas (rain) used as external validation datasets.

\subsection{Qualitative Results}

\subsubsection*{Across Sensors}
Qualitative results are generated using the physics-simulated VoxelScape dataset. While based on physical LiDAR modeling, the simulated intensities follow general physical behavior but lack sensor-specific response characteristics, resulting in structurally correct yet sensor-agnostic outputs.

Using our framework, we transfer sensor-specific intensity characteristics onto these scans by adapting them to real sensor domains such as SemanticKITTI and nuScenes. The scene geometry remains unchanged, while intensity patterns are modified according to the target sensor. As shown in Fig.~\ref{fig:gi_kitti} and Fig.~\ref{fig:gi_nuScenes}, the framework produces distinct, realistic sensor-aligned intensity distributions from a common simulated dataset.

\subsubsection*{Snow Adaptation}

Figure~\ref{fig:kitti_snow} shows qualitative results where clear-weather SemanticKITTI scans are transformed into snow conditions using the proposed framework. The method introduces snow-specific geometric and radiometric effects through physics modeling, followed by learning-based intensity refinement. The results exhibit characteristic snow signatures, including attenuation, reduced intensity, and localized dropouts, consistent with real observations. These outputs remain physically plausible while preserving the underlying scene structure.

\subsubsection*{Rain Adaptation}


\begin{table}[htbp]
\centering
\caption{Training Data Combinations for Snow Condition}
\label{tab:snow_training_data}
\resizebox{\columnwidth}{!}{%
\begin{tabular}{llll}
\toprule
\textbf{Condition} & \textbf{Method} & \textbf{Training Data} & \textbf{Total Frames} \\
\midrule
\multirow{5}{*}{\textbf{Snow}} 
  & \multirow{3}{*}{\textbf{Baseline}} 
      & KITTI-Clear & 4000 \\
  & 
      & KITTI-Clear + CADC & 2000 + 2000 \\
  & 
      & CADC (Real Snow) & 4000 \\
\cline{2-4}
  & \multirow{2}{*}{\textbf{Proposed}} 
      & KITTI-Snow (Ours) & 4000 \\
  & 
      & KITTI-Snow (Ours) + CADC & 2000 + 2000 \\
\midrule
\multirow{5}{*}{\textbf{Snow}} 
  & \multirow{3}{*}{\textbf{Baseline}} 
      & nuScenes-Clear & 4000 \\
  & 
      & nuScenes-Clear + CADC & 2000 + 2000 \\
  & 
      & CADC (Real Snow) & 4000 \\
\cline{2-4}
  & \multirow{2}{*}{\textbf{Proposed}} 
      & nuScenes-Snow (Ours) & 4000 \\
  & 
      & nuScenes-Snow (Ours) + CADC & 2000 + 2000 \\
\bottomrule
\end{tabular}%
}
\end{table}


\begin{table*}[ht]
\centering
\caption{PV-RCNN object detection performance under snow conditions using different training dataset combinations for KITTI and nuScenes. Synthetic snow generated by our ReaLiTy framework significantly improves robustness, especially when combined with real CADC snow data.}
\renewcommand{\arraystretch}{1.2}
\resizebox{\linewidth}{!}{
\begin{tabular}{llcccccc}
\toprule
\textbf{Class} & \textbf{Training Data} 
& \textbf{3D (0.7)} & \textbf{3D (0.5)} 
& \textbf{BEV (0.7)} & \textbf{BEV (0.5)} 
& \textbf{Notes} \\
\midrule

\multicolumn{7}{l}{\textbf{KITTI $\rightarrow$ Snow Evaluation (Tested on CADC Real Snow)}} \\
\midrule
Car 
& KITTI-Clear 
& 6.8  & 14.9 & 7.9  & 18.4 & Poor snow generalization \\
& KITTI-Snow (Ours) 
& 28.4 & 36.1 & 30.2 & 41.8 & Synthetic snow improves robustness \\
& KITTI-Clear + CADC 
& 32.6 & 39.4 & 34.1 & 45.8 & Real snow data improves robustness \\
& KITTI-Snow (Ours) + CADC 
& 39.5 & 46.2 & 41.8 & 49.9 & Best synthetic+real combination \\
& CADC (Real Snow) 
& 47.9 & 53.8 & 49.6 & 56.9 & Real-snow upper bound \\

\midrule
Pedestrian
& KITTI-Clear 
& 11.2 & 19.3 & 12.1 & 21.4 & Poor snow generalization \\
& KITTI-Snow (Ours) 
& 25.8 & 31.4 & 27.3 & 34.8 & Synthetic snow improves robustness \\
& KITTI-Clear + CADC 
& 27.4 & 33.9 & 29.1 & 37.8 & Real snow data improves robustness \\
& KITTI-Snow (Ours) + CADC 
& 34.9 & 40.1 & 36.8 & 43.7 & Best synthetic+real combination \\
& CADC (Real Snow) 
& 41.6 & 47.5 & 43.9 & 49.8 & Real-snow upper bound \\

\midrule
Truck
& KITTI-Clear 
& 4.5  & 10.6 & 5.8  & 12.4 & Poor snow generalization \\
& KITTI-Snow (Ours) 
& 27.4 & 33.4 & 29.1 & 36.2 & Synthetic snow improves robustness \\
& KITTI-Clear + CADC 
& 30.1 & 36.5 & 32.8 & 39.4 & Real snow data improves robustness \\
& KITTI-Snow (Ours) + CADC 
& 37.8 & 42.1 & 40.2 & 45.6 & Best synthetic+real combination \\
& CADC (Real Snow) 
& 44.3 & 49.8 & 46.9 & 52.1 & Real-snow upper bound \\

\midrule
\multicolumn{7}{l}{\textbf{nuScenes $\rightarrow$ Snow Evaluation (Tested on CADC Real Snow)}} \\
\midrule
Car 
& nuScenes-Clear 
& 5.9  & 12.8 & 7.1  & 16.4 & Poor snow generalization \\
& nuScenes-Snow (Ours) 
& 26.4 & 33.2 & 28.1 & 40.2 & Synthetic snow improves robustness \\
& nuScenes-Clear + CADC 
& 29.1 & 36.4 & 30.7 & 43.2 & Real snow data improves robustness \\
& nuScenes-Snow (Ours) + CADC 
& 36.7 & 44.1 & 39.2 & 47.3 & Best synthetic+real combination \\
& CADC (Real Snow)
& 47.9 & 53.8 & 49.6 & 56.9 & Real-snow upper bound \\

\midrule
Pedestrian
& nuScenes-Clear 
& 9.1  & 16.4 & 10.7 & 19.2 & Poor snow generalization \\
& nuScenes-Snow (Ours) 
& 24.1 & 29.8 & 26.5 & 33.6 & Synthetic snow improves robustness \\
& nuScenes-Clear + CADC 
& 25.9 & 32.1 & 27.4 & 36.5 & Real snow data improves robustness \\
& nuScenes-Snow (Ours) + CADC 
& 33.4 & 39.1 & 35.8 & 42.3 & Best synthetic+real combination \\
& CADC (Real Snow) 
& 41.6 & 47.5 & 43.9 & 49.8 & Real-snow upper bound \\

\midrule
Truck
& nuScenes-Clear 
& 3.8  & 9.2  & 4.9  & 11.8 & Poor snow generalization \\
& nuScenes-Snow (Ours) 
& 25.8 & 31.2 & 27.6 & 34.5 & Synthetic snow improves robustness \\
& nuScenes-Clear + CADC 
& 27.2 & 33.4 & 29.1 & 38.6 & Real snow data improves robustness \\
& nuScenes-Snow (Ours) + CADC 
& 35.4 & 41.6 & 37.9 & 44.1 & Best synthetic+real combination \\
& CADC (Real Snow) 
& 44.3 & 49.8 & 46.9 & 52.1 & Real-snow upper bound \\

\bottomrule
\end{tabular}
}
\label{tab:od-snow}
\end{table*}


Figure~\ref{fig:nuscenes_rain} shows qualitative results where clear-weather nuScenes LiDAR scans are transformed into rain conditions using the same pipeline. The framework introduces rain-induced effects while preserving the underlying scene layout. The results capture characteristic rain signatures such as distance-dependent attenuation and mild backscatter-induced intensity elevation. Compared with real Boreas rain data, the adapted scans exhibit similar degradation patterns, indicating realistic paired clear-to-rain LiDAR observations.

\subsection{Downstream Evaluation: 3D Object Detection}

To assess the practical value of the proposed weather–adapted LiDAR data, we evaluate its impact on a downstream perception task. We train PV-RCNN, a state-of-the-art 3D object detector~\cite{pvrcnn_shi2020pv}, implemented using the OpenPCDet framework~\cite{openpcdet2020}, on multiple training configurations involving \emph{clear}, \emph{real snow}, and \emph{simulated snow} point clouds. This controlled setup enables us to quantify how well the snow-augmented point clouds generated by the ReaLiTy framework improve detection robustness under adverse conditions, and whether they provide complementary benefits to limited real-weather datasets.

\subsubsection{Training Data Combinations}

Table~\ref{tab:snow_training_data} summarizes all data configurations used for training the object detector under snow conditions. Clear-weather KITTI and nuScenes scans serve as the reference inputs, and the corresponding snow-adapted scans generated by ReaLiTy (KITTI-Snow and nuScenes-Snow) form the proposed paired data. CADC snow scans are used as the real-world benchmark for comparison under snow conditions. All datasets are captured using automotive-grade multi-beam LiDAR sensors, and the evaluation focuses on comparing weather-induced geometric and intensity characteristics across clear, generated-snow, and real-snow observations.

\subsubsection{Evaluation Protocol}

The PVRCNN detector is trained using the data combinations described above and evaluated \emph{exclusively} on the CADC snow dataset.  
This ensures a realistic assessment of generalization to severe winter conditions.

The evaluation metric is class-wise Average Precision (AP) for cars, pedestrians, and trucks, reported for both 3D and BEV detection at standard IoU thresholds (0.7 and 0.5)

The trends observed in Table~\ref{tab:od-snow} show consistent relationships between training data composition and downstream 3D detection performance under snow conditions. The behaviour across dataset combinations and object classes can be explained as follows:

\begin{itemize}

\item Training on clear-weather datasets alone (KITTI-Clear or nuScenes-Clear) yields very poor performance when evaluated on real CADC snow data across all classes and IoU thresholds. This is expected because clear-weather scans do not contain snow-induced attenuation, backscatter noise, and sparsity artifacts. As a result, the learned feature distributions do not match the snow domain.

\item Adding a subset of real CADC snow data to clear-weather training (Clear + CADC) produces a large performance jump for all classes. This improvement comes from direct exposure to true snow-induced intensity distortions and sampling patterns. However, the results remain below CADC-only training because half of the training data still comes from clear-weather distributions, which dilutes the weather-specific feature learning.

\item Training exclusively on CADC real snow data produces the strongest performance in the table and serves as the practical upper bound in this study, since both training and evaluation distributions are matched. This confirms that real adverse-weather measurements remain the most informative supervision source when available. All mixed and synthetic configurations approach this fully matched training setting.

\item Training on the proposed generated snow datasets alone (KITTI-Snow (Ours) or nuScenes-Snow (Ours)) consistently and substantially outperforms their corresponding clear-only baselines. This demonstrates that the proposed ReaLiTy framework successfully transfers snow-specific radiometric characteristics into clear-weather scans. The detector therefore learns weather-aware features without being explicitly trained on real snow scans.

\item Combining generated snow data with real CADC snow data (KITTI-Snow (Ours) + CADC and nuScenes-Snow (Ours) + CADC) consistently improves over the corresponding Clear + CADC combinations across all classes. This shows that the adapted synthetic data provides complementary diversity rather than redundant samples. The generated snow scans expand coverage of object pose, distance, and scene structure under snow-like intensity statistics, while the real CADC frames anchor the model to true sensor-weather responses. The two sources, therefore, act synergistically.

\end{itemize}

\section{Conclusion}
This work presented \textit{ReaLiTy}, a unified physics-informed learning framework for realistic LiDAR transformation and adverse-weather simulation. By combining physically grounded modeling with learning-based refinement, the framework reproduces sensor-specific intensity characteristics while capturing weather-induced degradations. Leveraging cues such as range, incidence angle, and reflectance ensures physically consistent outputs and reduces the gap between simulated and real-world LiDAR observations.

A key contribution is the release of the \textit{LiDAR Adaptation Dataset Suite (LADS)}, a collection of derived point cloud datasets generated using ReaLiTy. LADS provides physically consistent sensor and weather transformations in the native format of source datasets, enabling direct one-to-one comparisons. It includes paired data across sensor and weather domains, supporting benchmarking, cross-sensor transfer, adverse-weather robustness evaluation, and simulation-to-real validation. To our knowledge, LADS is the first dataset to systematically provide unified LiDAR transformations across multiple sensors and weather conditions.

Together, ReaLiTy and LADS enable reproducible research in LiDAR realism and adaptation. While this work primarily evaluates weather adaptation, sensor adaptation is demonstrated through dataset generation and qualitative analysis. The framework is extensible to new datasets, sensors, and weather conditions, and LADS will be expanded with additional paired data. Overall, this work provides a scalable foundation for improving LiDAR simulation and supporting robust perception in diverse operating conditions.

\section*{Acknowledgment}

We thank the UNB, Canada, for Compute Canada HPC access, DS Research (FFT Division, DST, GOI) and SimDaaS Autonomy Pvt. Ltd for support.

\bibliographystyle{ieeetr}
\bibliography{reference}

\begin{thebibliography}{10}

\bibitem{wang2018pointseg}
Y.~Wang, T.~Shi, P.~Yun, L.~Tai, and M.~Liu, ``Pointseg: Real-time semantic
  segmentation based on 3d lidar point cloud,'' {\em arXiv preprint
  arXiv:1807.06288}, 2018.

\bibitem{anand2024toward}
V.~Anand, B.~Lohani, G.~Pandey, and R.~Mishra, ``Toward physics-aware deep
  learning architectures for lidar intensity simulation,'' {\em arXiv preprint
  arXiv:2404.15774}, 2024.

\bibitem{anand2026siat}
V.~Anand, S.~Yadav, M.~Limba, G.~Pandey, {\em et~al.}, ``Realistic lidar data
  simulation for autonomous systems using physics-informed learning,'' in {\em
  SIAT}, January 2026.
\newblock SAE Technical Paper.

\bibitem{hahner2022lidar}
M.~Hahner, C.~Sakaridis, M.~Bijelic, F.~Heide, F.~Yu, D.~Dai, and L.~Van~Gool,
  ``Lidar snowfall simulation for robust 3d object detection,'' in {\em
  Proceedings of the IEEE/CVF CVPR}, pp.~16364--16374, 2022.

\bibitem{vacek2021learning}
P.~Vacek, O.~Ja{\v{s}}ek, K.~Zimmermann, and T.~Svoboda, ``Learning to predict
  lidar intensities,'' {\em IEEE Transactions on Intelligent Transportation
  Systems}, vol.~23, no.~4, pp.~3556--3564, 2021.

\bibitem{Vivek_Advancing}
V.~Anand, B.~Lohani, G.~Pandey, and R.~Mishra, ``Advancing lidar intensity
  simulation through learning with novel physics-based modalities,'' {\em IEEE
  Transactions on Intelligent Transportation Systems}, vol.~26, no.~5,
  pp.~6493--6502, 2025.

\bibitem{simdaas}
{SimDaaS}, ``{SimDaaS Simulator}.'' \url{https://simdaas.com/}, 2026.
\newblock Accessed: April 1, 2026.

\bibitem{manivasagam2020lidarsim}
S.~Manivasagam, S.~Wang, K.~Wong, W.~Zeng, M.~Sazanovich, S.~Tan, B.~Yang,
  W.-C. Ma, and R.~Urtasun, ``Lidarsim: Realistic lidar simulation by
  leveraging the real world,'' in {\em Proceedings of the IEEE/CVF Conference
  on Computer Vision and Pattern Recognition}, pp.~11167--11176, 2020.

\bibitem{guillard2022learning}
B.~Guillard, S.~Vemprala, J.~K. Gupta, O.~Miksik, V.~Vineet, P.~Fua, and
  A.~Kapoor, ``Learning to simulate realistic lidars,'' in {\em 2022 IEEE/RSJ
  International Conference on Intelligent Robots and Systems (IROS)},
  pp.~8173--8180, IEEE, 2022.

\bibitem{learning-based-review}
H.~Haghighi, X.~Wang, H.~Jing, and M.~Dianati, ``Review of the learning-based
  camera and lidar simulation methods for autonomous driving systems,'' {\em
  arXiv preprint arXiv:2402.10079}, 2024.

\bibitem{bijelic2020weather}
M.~Bijelic, T.~Gruber, and W.~Ritter, ``Seeing through fog without seeing fog:
  Deep multimodal sensor fusion in adverse weather,'' in {\em CVPR}, 2020.

\bibitem{zhang2023fogsimulation}
Z.~Zhang, Y.~Liu, {\em et~al.}, ``Fog simulation on real lidar point clouds for
  3d object detection in adverse weather,'' in {\em Proceedings of the IEEE/CVF
  International Conference on Computer Vision}, 2023.

\bibitem{fang2023lidardiffusion}
R.~Fang, B.~Yu, {\em et~al.}, ``Lidardiffusion: Learning realistic lidar
  simulation via diffusion models,'' in {\em NeurIPS}, 2023.

\bibitem{behley2019iccv}
J.~Behley, M.~Garbade, A.~Milioto, J.~Quenzel, S.~Behnke, C.~Stachniss, and
  J.~Gall, ``Semantickitti: A dataset for semantic scene understanding of lidar
  sequences,'' in {\em Proc. of the IEEE/CVF International Conf. on Computer
  Vision (ICCV)}, 2019.

\bibitem{caesar2020nuscenes}
H.~Caesar, V.~Bankiti, A.~H. Lang, S.~Vora, V.~E. Liong, Q.~Xu, A.~Krishnan,
  Y.~Pan, G.~Baldan, and O.~Beijbom, ``nuscenes: A multimodal dataset for
  autonomous driving,'' in {\em Proceedings of the IEEE/CVF conference on
  computer vision and pattern recognition}, pp.~11621--11631, 2020.

\bibitem{geiger2012we}
A.~Geiger, P.~Lenz, and R.~Urtasun, ``Are we ready for autonomous driving? the
  kitti vision benchmark suite,'' in {\em 2012 IEEE conference on computer
  vision and pattern recognition}, pp.~3354--3361, IEEE, 2012.

\bibitem{mclean2020cadc}
J.~McLean, Y.~Hegde, U.~Masood, A.~Bilal, R.~P. Wildes, and S.~L. Waslander,
  ``Canadian adverse driving conditions dataset,'' {\em arXiv preprint
  arXiv:2001.10117}, 2020.

\bibitem{lisa_kilic2021lidar}
V.~Kilic, D.~Hegde, V.~Sindagi, A.~B. Cooper, M.~A. Foster, and V.~M. Patel,
  ``Lidar light scattering augmentation (lisa): Physics-based simulation of
  adverse weather conditions for 3d object detection,'' {\em arXiv preprint
  arXiv:2107.07004}, 2021.

\bibitem{anand_snow_iv}
V.~Anand, B.~Lohani, R.~Mishra, and G.~Pandey, ``Towards realistic lidar
  intensity simulation in snowy weather using physics-informed learning,'' in
  {\em 2025 IEEE Intelligent Vehicles Symposium (IV)}, pp.~2552--2557, 2025.

\bibitem{anand2026sim2real}
V.~Anand, B.~Lohani, V.~Kumar, R.~Mishra, and G.~Pandey, ``Toward closing the
  sim-to-real gap: A physics-guided learning approach for lidar intensity
  simulation,'' {\em IEEE Transactions on Intelligent Transportation Systems},
  2026.
\newblock Early access.

\bibitem{anand2026sim2real_aw}
V.~Anand, B.~Lohani, R.~Mishra, and G.~Pandey, ``Simulating realistic lidar
  data under adverse weather for autonomous vehicles: A physics-informed
  learning approach,'' 2026.
\newblock arXiv preprint arXiv:2604.01254.

\bibitem{pvrcnn_shi2020pv}
S.~Shi, C.~Guo, L.~Jiang, Z.~Wang, J.~Shi, X.~Wang, and H.~Li, ``Pv-rcnn:
  Point-voxel feature set abstraction for 3d object detection,'' in {\em
  Proceedings of the IEEE/CVF CVPR}, pp.~10529--10538, 2020.

\bibitem{openpcdet2020}
O.~D. Team, ``Openpcdet: An open-source toolbox for 3d object detection from
  point clouds.'' \url{https://github.com/open-mmlab/OpenPCDet}, 2020.

\end{thebibliography}
\patchcmd{\thebibliography}{\baselineskip15pt}{\baselineskip 12pt}{}{}

\begin{IEEEbiography}[{\includegraphics[width=1in,height=1.25in,clip,keepaspectratio]{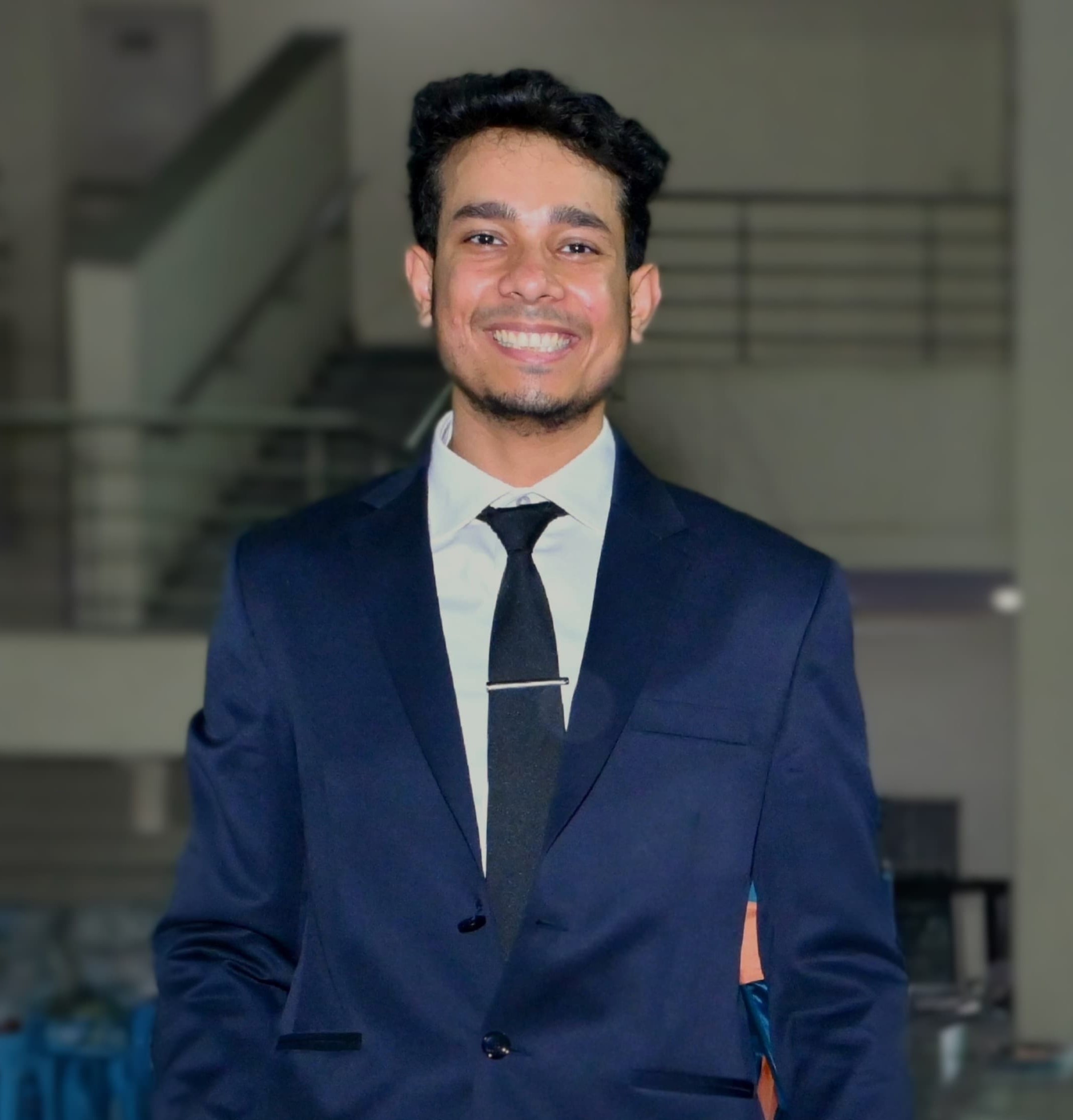}}]{Vivek Anand} 
received the Ph.D. degree in Geoinformatics from the Indian Institute of Technology (IIT) Kanpur in 2026. He is currently a Fellow of Academic and research Excellence at IIT Kanpur. His research interests include physics-informed LiDAR simulation, sim-to-real gap analysis, and autonomous systems. Postal Address: IIT Kanpur, Uttar Pradesh 208016, India. Email: viveka21@iitk.ac.in.
\end{IEEEbiography}

\begin{IEEEbiography}[{\includegraphics[width=1in,height=1.25in,clip,keepaspectratio]{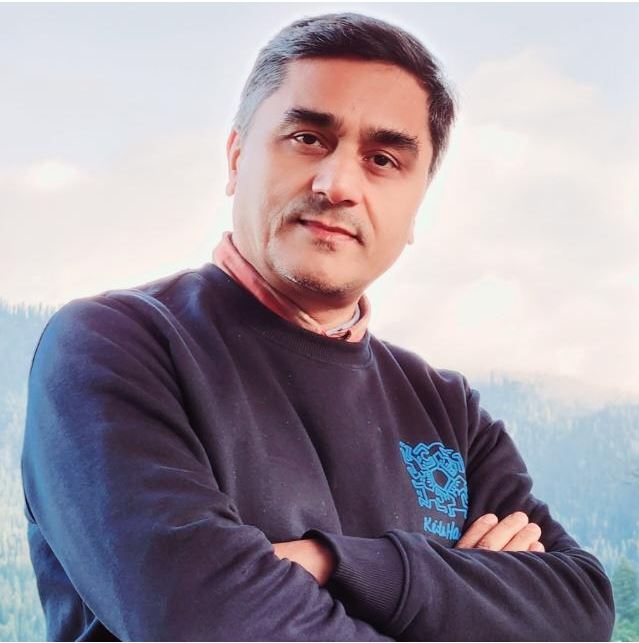}}]{Bharat Lohani} 
received the Ph.D. degree from the University of Reading, U.K., in 1999. He is currently a Professor with the Department of Civil Engineering, IIT Kanpur. His research interests include LiDAR simulation, point cloud classification using deep learning, and forest conservation. Postal Address: IIT Kanpur, Uttar Pradesh 208016, India. Email: blohani@iitk.ac.in.
\end{IEEEbiography}

\begin{IEEEbiography}[{\includegraphics[width=1in,height=1.25in,clip,keepaspectratio]{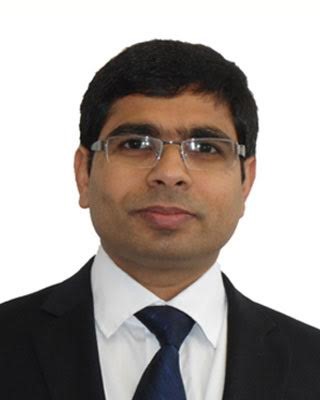}}]{Rakesh Mishra} 
received the Ph.D. degree from the University of New Brunswick, Canada, in 2017. He is currently an Adjunct Professor with the GGE Department, University of New Brunswick, and CTO of SceneSharp Technologies Inc. His interests include image fusion and LiDAR. Postal Address: University of New Brunswick, NB, Canada. Email: rakesh.mishra@unb.ca.
\end{IEEEbiography}

\begin{IEEEbiography}[{\includegraphics[width=1in,height=1.25in,clip,keepaspectratio]{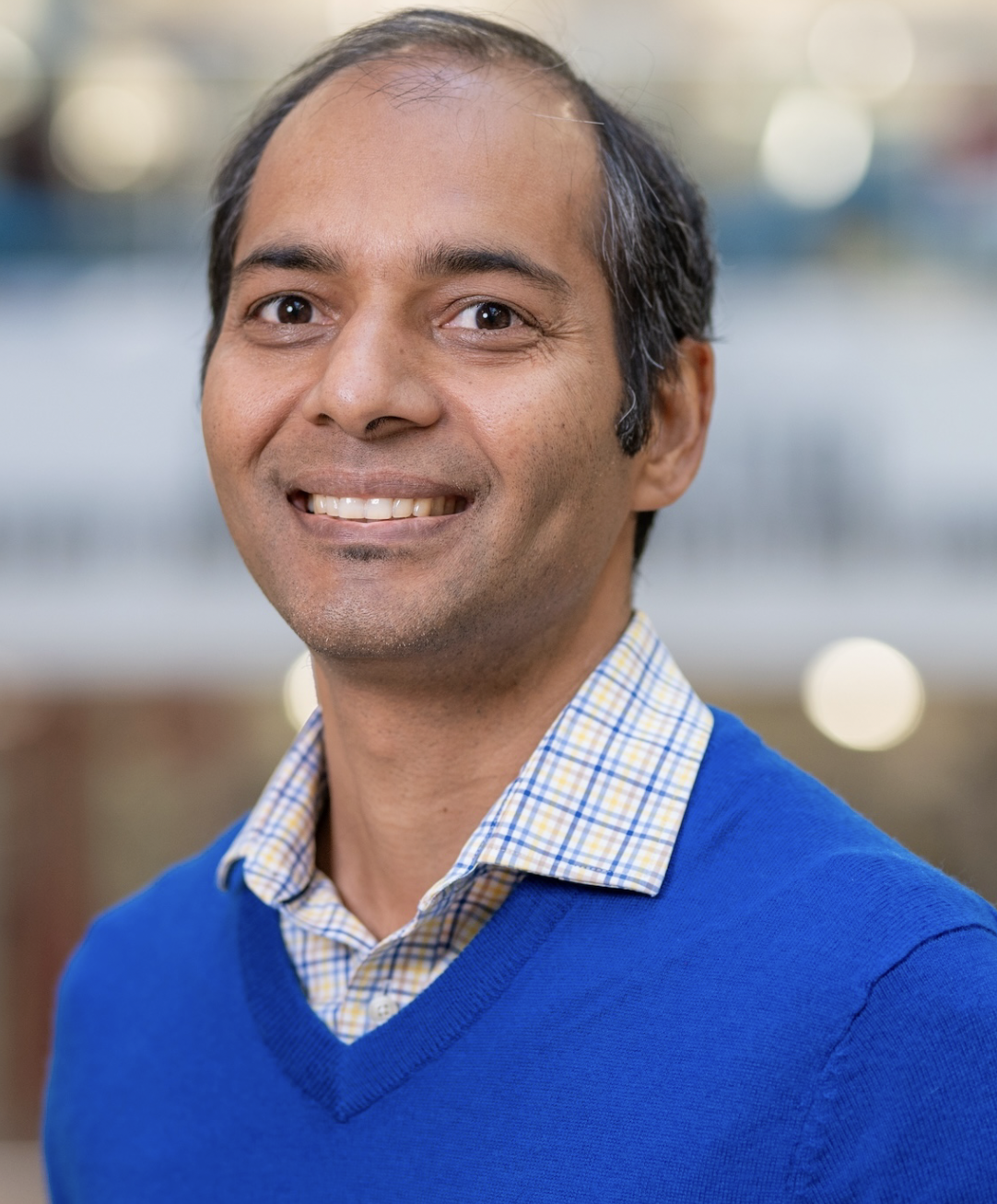}}]{Gaurav Pandey} 
received the Ph.D. degree from the University of Michigan, Ann Arbor, in 2013. He is currently an Associate Professor with the ETID department, Texas A\&M University. His research focuses on autonomous vehicles and robotics. Postal Address: Texas A\&M University, College Station, TX, USA. Email: gaurav.pandey@tamu.edu.
\end{IEEEbiography}

\end{document}